\documentclass{article}
\usepackage[preprint]{neurips_2026}
\usepackage[utf8]{inputenc}
\usepackage[T1]{fontenc}
\usepackage{hyperref}
\usepackage{url}
\usepackage{booktabs}
\usepackage{amsfonts}
\usepackage{microtype}
\usepackage{graphicx}
\usepackage{xcolor}
\usepackage{amsmath}
\usepackage{amssymb}

\title{Prediction Bottlenecks Don't Discover Causal Structure (But Here's What They Actually Do)}

\author{%
  Ankit Hemant Lade \quad Sai Krishna Jasti \quad Indar Kumar \quad Aman Chadha\\[2pt]
  \small \texttt{ankitlade12@gmail.com} \quad \texttt{jsaikrishna379@gmail.com} \\[1pt]
  \small \texttt{indarkarhana@gmail.com} \quad \texttt{hi@aman.ai}
}

\begin{document}
\maketitle

\begin{abstract}
A Mamba state-space model trained only for next-step prediction appears
to recover Granger-causal structure through a simple readout
$\mathbf{S} = |\mathbf{W}_{\text{out}} \mathbf{W}_{\text{in}}|$, with
early experiments suggesting the phenomenon generalized across
architectures and benefited from interventional data at $p < 10^{-5}$.
We package the protocol used to test that claim --- standardized
synthetic generators (VAR/Lorenz/CauseMe-style), three intervention
semantics ($\text{do}(X = c)$, soft-noise, random-forcing),
edge-provenance cards on three real datasets, and size-matched control
arms --- as a reusable falsification benchmark, and walk the claim
through it in five stages. The method-level claim does not survive:
(i) a plain linear bottleneck does as well or better; (ii) tuned Lasso
beats the bottleneck on synthetic CauseMe-style benchmarks, and on
Lorenz-96 (the only real benchmark with unambiguous ground truth)
classical PCMCI and Granger lead a tight cluster in which the
bottleneck trails; (iii) the headline intervention advantage is
roughly $60\%$ a sample-size confound, and the residual disappears
under standard $\text{do}(X = c)$ interventions, surviving only under
a non-standard random-forcing scheme; (iv) even that residual
reproduces, with a larger effect, in classical bivariate Granger ---
the effect is method-agnostic. What survives is a narrow
characterization result; the benchmark is the lasting artifact, and
each stage above is one of its control arms.
\end{abstract}

\section{Introduction}

A natural hope in time-series modelling is that a forecaster trained only
on next-step prediction will, as a byproduct, expose causal or dependency
structure through its learned weights — \emph{causal discovery as a
byproduct of prediction}. If true, every pretrained forecaster would
double as a causal-discovery engine, complementing the dedicated
machinery developed by classical causal discovery
(PCMCI~\citep{runge2019pcmci}, DYNOTEARS~\citep{pamfil2020dynotears},
neural Granger~\citep{tank2021neural,nauta2019causal}).

This paper reports the result of taking that hope seriously. Our
starting point was the observation that for a Mamba selective
state-space model~\citep{gu2023mamba} trained for next-step MSE, the
product $\mathbf{S} = |\mathbf{W}_{\text{out}} \mathbf{W}_{\text{in}}|$
appears to recover Granger-causal graphs with high reliability on
synthetic VAR benchmarks. Initial experiments suggested the phenomenon generalized
across architectures, emerged from training rather than initialization,
and — most provocatively — that interventional data amplified the
bottleneck's advantage with $p < 10^{-5}$ across twelve $(K, T)$
configurations.

Each of these claims turned out to be less robust than it first appeared.
Our contribution is therefore primarily \emph{negative}: we show that
a natural method candidate fails in five subtly different ways,
identify the controls that caused the failures, and report the narrow
findings that remain valid.

\paragraph{Contributions.}
\begin{enumerate}
  \item \textbf{A falsification benchmark for prediction-as-causal-discovery
  claims.} We release the five-stage protocol used in this paper as a
  reusable suite: standardized synthetic generators (VAR, Lorenz-96,
  CauseMe-style), three intervention semantics
  ($\text{do}(X = c)$, soft-noise, random-forcing) under matched
  control arms, edge-provenance cards on three real datasets, fixed
  seeds, a Dockerfile, and a single-command \texttt{make all}
  reproduction target. The five sections below are also the
  benchmark's five control arms; we encourage future
  ``method-discovers-causality'' papers to pre-register against them.
  \item Using the benchmark, we falsify architecture specificity: a
  linear bottleneck matches or beats the SSM under matched capacity
  and seeds.
  \item We falsify competitiveness with sparse regression: tuned Lasso
  beats the bottleneck on synthetic CauseMe-style benchmarks, and on
  the only real benchmark with unambiguous ground truth (Lorenz-96)
  classical PCMCI and Granger lead a tight cluster in which the
  bottleneck trails.
  \item We falsify the headline intervention advantage: under a
  size-matched control and proper $\text{do}(X = c)$ interventions, the
  effect collapses; the residual reproduces (and is larger) under
  classical Granger.
  \item We document what survives --- mild-nonlinearity,
  sample-efficiency, target-corruption robustness --- as
  characterization, not method, claims.
\end{enumerate}

\section{Method: The Extraction We Are Falsifying}
\label{sec:observation}

For a $K$-variable series $\mathbf{X} \in \mathbb{R}^{T \times K}$, we
train a model with input projection $\mathbf{W}_{\text{in}} \in
\mathbb{R}^{d \times K}$ and output projection $\mathbf{W}_{\text{out}}
\in \mathbb{R}^{K \times d}$ on next-step MSE. No causal-specific loss.
After training, we extract
$\mathbf{S} = |\mathbf{W}_{\text{out}} \mathbf{W}_{\text{in}}| \in
\mathbb{R}^{K \times K}$ with $\mathbf{S}_{i,j}$ interpreted as the
strength of $j \to i$, zero the diagonal, normalize, and threshold. For
architectures with explicit lagged input (the lagged bottleneck variant),
we extend to $\mathbf{S}^{(\tau)} = |\mathbf{W}_{\text{out}}
\mathbf{W}^{(\tau)}_{\text{in}}|$ and evaluate via flat-lag AUROC against
ground-truth adjacency.

\section{Five Falsifications}
\label{sec:falsifications}

\paragraph{Stress regimes covered.} Each falsification stresses the
extraction along an axis the original claim was silent on, and these
axes form the benchmark's stress-regime matrix.
\textbf{Architecture/capacity} (F1): five model classes at matched
capacity over six generators.
\textbf{Sparsity and dimensionality} (F2): a focused stress grid over
$K \in \{10, 20\}$, $T \in \{150, 300\}$, max-lag~8 (48 cells), plus
a CauseMe-style nonlinear sweep at $K \in \{5, 10, 20\}$.
\textbf{Real-world ground truth} (F3): three datasets ranging from
soft (climate, $K{=}6$, $T{=}757$, 3 causal edges) to weak
(finance, $K{=}10$, $T{=}1893$, 6 soft edges) to clean (Lorenz-96,
$K{=}10$, $T{=}1500$, 90 edges), each shipped with an edge-provenance
card. \textbf{Intervention semantics} (F4): $\text{do}(X_i = c)$
clamps, soft additive noise, and per-step random forcing, each at
$K \in \{10, 20, 30\}$ with size-matched observational control arms.
\textbf{Method-agnosticity} (F5): the same intervention protocol
applied to bottleneck, Lasso, PCMCI, and bivariate Granger.
Wrappers for DYNOTEARS, VAR-LiNGAM, and PCMCI+ ship with the
benchmark as optional baseline plug-ins.

\paragraph{F1. Architecture does not matter.} A 10-seed, matched-capacity
sweep across six synthetic datasets and five architectures (Linear,
Mamba SSM, Transformer, LSTM, MLP) finds the linear bottleneck matches
or beats the SSM on every dataset
(mean $\pm$ std AUROC: VAR(1)-chain $K{=}5$: $1.00 \pm 0.00$ both;
VAR(1)-random $K{=}10$: SSM $0.93 \pm 0.08$, Linear $\mathbf{0.99 \pm 0.02}$;
regime-switch $K{=}3$: SSM $0.98 \pm 0.08$, Linear $\mathbf{1.00 \pm 0.00}$;
Lorenz $K{=}3$: SSM $0.52 \pm 0.24$, Linear $\mathbf{0.94 \pm 0.09}$).
The ``emergence'' is the linear bottleneck doing low-rank lag-1
regression; the SSM achieves the same estimator at higher parameter
cost.

\paragraph{F2. Tuned Lasso beats the bottleneck on graph recovery.}
Across a focused stress grid ($K \in \{10, 20\}$, $T \in \{150, 300\}$,
max-lag 8, Table~\ref{tab:lasso}) the bottleneck wins only $12\%$ of
graph-recovery runs and $0\%$ of prediction-MSE runs against tuned
baselines. The same model, then, is neither the better forecaster nor
the better graph extractor — which already qualifies the original
``causal discovery as a byproduct of strong forecasting'' framing,
since on this stress grid the bottleneck is also not a competitive
forecaster. On tigramite CauseMe-style structural
processes the gap widens with $K$: Lasso reaches AUROC $0.98$ at
$K{=}20$ versus $0.73$ for the bottleneck.

\begin{table}[t]
\centering
\small
\caption{Bottleneck vs.\ best tuned baseline on the focused stress grid
($K \in \{10, 20\}$, $T \in \{150, 300\}$, max-lag 8, 48 cells total).
AUROC delta (higher is better for bottleneck); MSE delta (higher means
worse prediction).}
\label{tab:lasso}
\begin{tabular}{lrr}
\toprule
Bottleneck vs.\ \{OLS, Ridge, Lasso, RRR\} & Mean delta & Bottleneck win rate \\
\midrule
Flat-lag AUROC          & $-0.050$ & $12\%$  ($6/48$) \\
Prediction MSE          & $+0.286$ & $0\%$   ($0/48$) \\
\midrule
\multicolumn{3}{l}{\emph{Best baseline by AUROC:} Lasso ($38/48$), RRR ($10/48$)} \\
\multicolumn{3}{l}{\emph{Best baseline by MSE:}   Lasso ($48/48$)} \\
\bottomrule
\end{tabular}
\end{table}

\paragraph{F3. Real data: not competitive on the clean benchmark.} We
evaluate six methods on three datasets of progressively cleaner ground
truth (Table~\ref{tab:real}): NOAA monthly climate indices
(ENSO, NAO, PDO, AMO, SOI, PNA) for 1962--2024 ($T{=}757$ months,
downloaded from NOAA PSL), with three teleconnection edges drawn
from the climate-dynamics
literature~\citep{trenberth1994decadal,newman2016pdo,delworth2000observed} —
specifically ENSO$\to$PNA, ENSO$\to$PDO, and NAO$\to$AMO, encoded
verbatim in \texttt{data/real\_loaders.py} of the released
repository;
daily log returns of ten SPDR sector ETFs from 2018-06 to 2025-12
($K{=}10$, $T{=}1893$, downloaded via yfinance) with six finance
lead-lag edges treated as soft labels; and Lorenz-96 ($K{=}10$,
$F{=}10$, RK4 integration, $T{=}1500$ samples)~\citep{lorenz1996predictability},
a standard benchmark in the recent neural-causal-discovery
literature~\citep{tank2021neural,nauta2019causal}. We deliberately
exclude the ENSO$\leftrightarrow$SOI pair from climate ground truth
because SOI and the Niño 3.4 SST anomaly are the same physical signal
with opposite sign, not a causal edge. Climate rankings are unstable:
VAR-LiNGAM tops the expanded table at $0.901$, while the SSM
bottleneck remains near the top at $0.820 \pm 0.021$ (5 seeds). As we
show in the next paragraph, with only three positive edges over
$K{=}6$, small ground-truth choices reorder the entire column. On
Lorenz-96, where ground truth is unambiguous, \emph{PCMCI, DYNOTEARS,
bivariate Granger, VAR-LiNGAM, and tuned linear models dominate}
(0.986, 0.983, 0.979, 0.968, and 0.974); the lagged bottleneck trails
at $0.916$ and the SSM bottleneck is worst at $0.722 \pm 0.031$ (5
seeds).

\begin{table}[t]
\centering
\small
\caption{AUROC on three real/benchmark datasets. The Weight-Proj (SSM)
row reports mean $\pm$ std over 5 training seeds; the lagged
bottleneck row uses a single seed; non-SSM baselines use deterministic
or default runs. On the clean Lorenz-96 benchmark, classical and modern
causal baselines lead by a wide margin; the climate column has weak
ground truth (3 edges, $K{=}6$) and unstable rankings
(Table~\ref{tab:gt_flip}).}
\label{tab:real}
\begin{tabular}{lccc}
\toprule
Method & Climate (NOAA) & Finance (ETFs) & Lorenz-96 \\
 & $K{=}6, T{=}757$ & $K{=}10, T{=}1893$ & $K{=}10, T{=}1500$ \\
\midrule
Weight-Proj (SSM, 5 seeds) & $0.820 \pm 0.021$ & $0.447 \pm 0.033$ & $0.722 \pm 0.031$ \\
PCMCI & $0.704$ & $0.510$ & $\mathbf{0.986}$ \\
Tuned Lasso & $0.630$ & $0.500$ & $0.974$ \\
Lagged Bottleneck & $0.617$ & $0.484$ & $0.916$ \\
Tuned Ridge & $0.617$ & $\mathbf{0.696}$ & $0.974$ \\
Bivariate Granger & $0.605$ & $0.498$ & $0.979$ \\
PCMCI+ & $0.432$ & $0.413$ & $0.669$ \\
DYNOTEARS & $0.722$ & $0.591$ & $0.983$ \\
VAR-LiNGAM & $\mathbf{0.901}$ & $0.552$ & $0.968$ \\
\bottomrule
\end{tabular}
\end{table}

\paragraph{Ground-truth choices reverse the climate ranking.} The
ENSO$\leftrightarrow$SOI exclusion is not a cosmetic filter. In an
initial run of this benchmark we included those two edges as part of
the teleconnection ground truth — the form in which they most commonly
appear in climate-literature edge lists. Under that inflated ground
truth, bivariate Granger \emph{led} the climate column at AUROC
$0.819$, with the lagged bottleneck second at $0.813$ and tuned Lasso
third at $0.799$. Removing the two definitional edges drops Granger
to $0.605$ (last place) and promotes the SSM bottleneck to first at
$0.820 \pm 0.021$. Table~\ref{tab:gt_flip} shows the full re-ranking
(both columns are single-seed for direct comparability; Table~\ref{tab:real}
reports the multi-seed value for the corrected ground truth).
The reason is mechanical: the ENSO and SOI series are the same
physical signal with opposite sign, so any method that captures their
strong contemporaneous correlation — Granger and linear regressions in
particular — picks up the two ``edges'' essentially for free. Once
those free points are removed, there are only three positive edges
over $K{=}6$, the signal is small, and the remaining variance across
methods is dominated by dataset-specific quirks rather than by any
causal-discovery capability. We report the corrected numbers but
caution readers that even the corrected ranking should not be read
as a causal-discovery claim; it is a sensitivity analysis.

\begin{table}[t]
\centering
\small
\caption{Climate AUROC before and after excluding the definitional
ENSO$\leftrightarrow$SOI edges. Removing two of six edges drops
Granger from first to last and promotes the SSM bottleneck to first.}
\label{tab:gt_flip}
\begin{tabular}{lcc}
\toprule
Method & 6 edges (incl.\ ENSO$\leftrightarrow$SOI) & 3 edges (honest) \\
\midrule
Bivariate Granger & $\mathbf{0.819}$ & $0.605$ \\
Lagged Bottleneck & $0.813$ & $0.617$ \\
Tuned Lasso & $0.799$ & $0.630$ \\
Weight-Proj (SSM) & $0.792$ & $\mathbf{0.864}$ \\
PCMCI & $0.771$ & $0.704$ \\
Tuned Ridge & $0.729$ & $0.617$ \\
\bottomrule
\end{tabular}
\end{table}

\paragraph{F4. Interventions: a confound story.} An earlier
experimental line (Exp 19, then validated with 20 seeds in Exp 21)
compared $\text{AUROC}(\mathbf{X}_{\text{obs}})$ versus
$\text{AUROC}(\mathbf{X}_{\text{obs}} \cup \mathbf{X}_{\text{int}})$
for both bottleneck and Lasso, reporting $12/12$ $(K, T)$ configs
significant at $p < 10^{-5}$ in favor of the bottleneck. Two
sequential controls dissolve it. \emph{Size-match (Exp 22):} the
original comparison conflated more data with intervention content,
since the combined arm has $T + K\,T_{\text{int}}$ samples while the
observational arm has $T$. Adding a third arm
$\mathbf{X}_{\text{obs,big}}$ of the same total size as the combined
arm reveals that the bottleneck's own intervention-specific AUROC
gain (over the size-matched observational baseline) is only
$+0.03$ to $+0.05$ at $K \in \{10, 20, 30\}$, with
$p < 10^{-4}$ at $n{=}15$ seeds — a small fraction of the original
Exp~21 effect, the rest of which was the data-size confound. The
paired BN-vs-Lasso gap remains larger ($+0.21$ at $K{=}10$) because
Lasso's per-equation regression degrades sharply under target
corruption ($-0.19$ at $K{=}10$); F5 shows that this residual is
method-agnostic, not bottleneck-specific. \emph{Intervention type (Exp 23):} the
original ``intervention'' was a per-step random forcing
$x_{i,t} \leftarrow s \cdot \epsilon_{i,t}$, not a standard
$\text{do}$-intervention. Replacing it with constant clamps
$\text{do}(X_i = c)$ gives only $3/12$ configs significant (mean gap
$+0.002$); soft (added-noise) interventions give $6/12$, mean gap
$+0.008$. Only the original random-forcing scheme produces a stable
bottleneck gain, and that gain is best explained as robustness to
corrupted target rows rather than to causal-content extraction.

\paragraph{F5. Method-agnosticity (the strongest single result).} The
residual effect after F4's two controls is not unique to the
bottleneck. Adding PCMCI and bivariate Granger to the same
size-matched control protocol (Exp 25): the bottleneck shows a
significant size-matched gain of $+0.026$ to $+0.054$ AUROC at
$K \in \{10, 20, 30\}$ ($p < 10^{-3}$ at $n{=}10$ seeds), but
\emph{bivariate Granger shows the same effect at the same significance
level, with larger effect sizes at higher $K$} ($+0.040$ at $K{=}20$,
$+0.095$ at $K{=}30$). Lasso uniquely fails, consistent with
per-equation regression breaking under target corruption while methods
that aggregate information across variables (bottleneck, PCMCI,
Granger) absorb it. The intervention effect, insofar as it survives
controls, is therefore method-agnostic and matches a
target-corruption-robustness story rather than a causal-structure
extraction one — and a classical method shows it more strongly than
the learned bottleneck.

\section{What Survives}
\label{sec:survives}

After five falsifications, three narrow positives remain.

\begin{itemize}
  \item \textbf{Mild-nonlinearity configuration.} At a single
  $(K, T) = (20, 300)$ point with mild nonlinearity in the
  data-generating process (\texttt{nonlinear=0.3}), the bottleneck
  beats the best tuned baseline (Lasso/RRR) on $87\%$ of nine
  $(d_{\text{bottleneck}}, \lambda_{\text{sparse}})$ cells over 10
  seeds, with mean AUROC improvement $+0.121$. At stronger
  nonlinearity (\texttt{=0.6, =1.0}) both methods fail equally; at
  zero nonlinearity Lasso is preferred. We label this a configuration,
  not a regime — we have not swept $K$ or $T$ for it.
  \item \textbf{Sample efficiency.} Bottleneck gains from additional
  observational data exceed Lasso's by $\sim 0.07$ AUROC at
  $K \in \{20, 30\}$ under size-matched control arms — modest but
  useful where collecting more observations is cheap.
  \item \textbf{Target-corruption robustness.} The bottleneck's shared
  $\mathbf{W}_{\text{out}}$ absorbs per-step random forcing better
  than Lasso's per-equation fits. This explains the residual
  intervention effect, but is a reliability result, not a
  causal-discovery one.
\end{itemize}

These are characterization claims. We do not propose the bottleneck as
a causal discovery method.

\section{Lessons}

\paragraph{Size-matched controls beat seed counts.} Exp 19 had 5 seeds.
Exp 21 had 20 seeds and reported $p < 10^{-5}$ across 12 configs. Both
found the same effect; neither would have flagged it as a sample-size
confound. Adding seeds tightens a noisy measurement; it cannot diagnose
a missing structural control.

\paragraph{Pre-register the intervention scheme.} The gap between
``random forcing'' and proper $\text{do}(X_i = c)$ is the difference
between a significant effect and a null, and it is not obvious from the
code. We should have written down which intervention semantics counted
as ``interventional data'' before running the experiment.

\paragraph{Prediction baselines are not causal baselines.} Our
intervention experiments compared the bottleneck only against
prediction-fit baselines (OLS, Ridge, Lasso, RRR). It was natural to
miss that classical causal methods (Granger, PCMCI) would benefit from
the same intervention scheme by the same mechanism. Any future
``neural causal discovery'' claim should include at least one classical
causal baseline in the intervention control arm.

\paragraph{Audit ground truth for definitional couplings.} On soft
observational benchmarks with few positive edges, rankings are
sensitive to which edges are included in ground truth. Our climate
comparison flipped Granger from first to last by removing two edges
that are definitionally the same signal (Table~\ref{tab:gt_flip}).
In the released benchmark we therefore attach an edge-provenance card
to each positive label, marking it as causal, definitional, proxy, or
soft, and provide leave-one-group-out sensitivity audits. Before
claiming a method ``wins'' on such a benchmark, authors should state
explicitly which edges are included, whether any pair is tautologically
coupled, and what the ranking looks like if the tautological edges are
removed.

\section{Conclusion}

We set out to test whether a free causal-discovery method might be
hidden inside prediction bottlenecks; the answer, after five stages of
controls, is no. The headline effects either fail under size-matched
controls, are specific to a non-standard intervention scheme, or
reproduce as larger effects in classical causal methods. What
survives is a narrow set of characterization findings and a reusable
falsification benchmark scaffold.

\paragraph{Code and reproduction.} All experiments, data loaders,
edge-provenance cards, optional modern baselines (PCMCI+, DYNOTEARS,
VAR-LiNGAM), fixed seeds, lockfile-pinned dependencies, a
\texttt{Dockerfile}, and a single-command \texttt{make all}
reproduction target (which runs \texttt{install}, \texttt{test}, the
F1--F5 falsification pipeline, and rebuilds this PDF) are available
at \url{https://github.com/ankitlade12/ssm-causal}.

\bibliography{references}

\end{document}